\pdfoutput=1

\documentclass[11pt]{article}

\usepackage{naacl2021}

\usepackage{times}
\usepackage{latexsym}

\usepackage[T1]{fontenc}

\usepackage[utf8]{inputenc}

\usepackage{microtype}

\usepackage{amsmath}
\usepackage{multirow}
\usepackage{graphicx}
\usepackage{amsfonts}
\usepackage{bm}
\usepackage{amssymb}
\usepackage{array}
\usepackage{color}
\usepackage{subfigure}

\newcommand{\todo}[1]{}
\renewcommand{\todo}[1]{{\color{red} TODO: {#1}}}

\newcommand{\tagforbt}{$ \langle \texttt{BT} \rangle $}
\newcommand{\fscore}{\ensuremath{\mathrm{F}_{0.5}}}

\newcommand{\uniform}{\textsc{Rep(Uni)}}
\newcommand{\parass}{\textsc{Rep(SS)}}
\newcommand{\similarity}{\textsc{Rep(Sim)}}
\newcommand{\worddrop}{\textsc{WDrop}}
\newcommand{\adv}{\textsc{Adv}}

\title{Rethinking Perturbations in Encoder-Decoders for Fast Training}

\author{Sho Takase \\
  Tokyo Institute of Technology \\
  \texttt{sho.takase@nlp.c.titech.ac.jp} \\\And
  Shun Kiyono \\
  RIKEN / Tohoku University \\
  \texttt{shun.kiyono@riken.jp} \\
 }

\begin{document}
\maketitle
\begin{abstract}
We often use perturbations to regularize neural models.
For neural encoder-decoders, previous studies applied the scheduled sampling~\cite{NIPS2015_5956} and adversarial perturbations~\cite{sato-etal-2019-effective} as perturbations but these methods require considerable computational time.
Thus, this study addresses the question of whether these approaches are efficient enough for training time.
We compare several perturbations in sequence-to-sequence problems with respect to computational time.
Experimental results show that the simple techniques such as word dropout~\cite{NIPS2016_6241} and random replacement of input tokens achieve comparable (or better) scores to the recently proposed perturbations, even though these simple methods are faster.
Our code is publicly available at \href{https://github.com/takase/rethink_perturbations}{https://github.com/takase/rethink\_perturbations}.
\end{abstract}

\section{Introduction}
Recent advances in neural encoder-decoders have driven tremendous success for sequence-to-sequence problems including machine translation~\cite{Sutskever:2014:SSL:2969033.2969173}, summarization~\cite{rush-chopra-weston:2015:EMNLP}, and grammatical error correction (GEC)~\cite{ji-etal-2017-nested}.
Since neural models can be too powerful, previous studies have proposed various regularization methods to avoid over-fitting.

To regularize neural models, we often apply a perturbation~\cite{Goodfellow2015,Miyato2017}, which is a small difference from a correct input.
During the training process, we force the model to output the correct labels for both perturbed inputs and unmodified inputs.
In sequence-to-sequence problems, existing studies regard the following as perturbed inputs: (1) sequences containing tokens replaced from correct ones~\cite{NIPS2015_5956,cheng-etal-2019-robust}, (2) embeddings injected small differences~\cite{sato-etal-2019-effective}.
For example, \newcite{NIPS2015_5956} proposed the scheduled sampling that samples a token from the output probability distribution of a decoder and uses it as a perturbed input for the decoder.
\newcite{sato-etal-2019-effective} applied an adversarial perturbation, which significantly increases the loss value of a model, to the embedding spaces of neural encoder-decoders.

Those studies reported that their methods are effective to construct robust encoder-decoders.
However, their methods are much slower than the training without using such perturbations because they require at least one forward computation to obtain the perturbation.
In fact, we need to run the decoder the same times as the required number of perturbations in the scheduled sampling~\cite{NIPS2015_5956}.
For adversarial perturbations~\cite{sato-etal-2019-effective}, we have to compute the backpropagation in addition to forward computation because we use gradients to obtain perturbations.

Those properties seriously affect the training budget.
For example, it costs approximately 1,800 USD for each run when we train Transformer (big) with adversarial perturbations~\cite{sato-etal-2019-effective} on the widely used WMT English-German training set in AWS EC2\footnote{We assume that we use on-demand instances having 8 V100 GPUs.}.
Most studies conduct multiple runs for the hyper-parameter search and/or model ensemble to achieve better performance~\cite{barrault-etal-2019-findings}, which incurs a tremendous amount of training budget for using such perturbations.
\newcite{strubell-etal-2019-energy} and \newcite{DBLP:journals/corr/abs-1907-10597} indicated that recent neural approaches increase computational costs substantially, and they encouraged exploring a cost-efficient method.
For instance, \newcite{li2020train} explored a training strategy to obtain the best model in a given training time.
However, previous studies have paid little attention to the costs of computing perturbations.

Thus, we rethink a time efficient perturbation method.
In other words, we address the question whether perturbations proposed by recent studies as effective methods are time efficient.
We compare several perturbation methods for neural encoder-decoders in terms of computational time.
We introduce light computation methods such as word dropout~\cite{NIPS2016_6241} and using randomly sampled tokens as perturbed inputs.
These methods are sometimes regarded as baseline methods~\cite{NIPS2015_5956}, but experiments on translation datasets indicate that these simple methods surprisingly achieve comparable scores to those of previous effective perturbations~\cite{NIPS2015_5956,sato-etal-2019-effective} in a shorter training time.
Moreover, we indicate that these simple methods are also effective for other sequence-to-sequence problems: GEC and summarization.

\section{Definition of Encoder-Decoder}
In this paper, we address sequence-to-sequence problems such as machine translation with neural encoder-decoders, and herein we provide a definition of encoder-decoders.

In sequence-to-sequence problems, neural encoder-decoders generate a sequence corresponding to an input sequence.
Let $\bm{x}_{1:I}$ and $\bm{y}_{1:J}$ be input and output token sequences whose lengths are $I$ and $J$, respectively: $\bm{x}_{1:I} = x_1, ..., x_I$ and $\bm{y}_{1:J} = y_1, ..., y_J$.
Neural encoder-decoders compute the following conditional probability:
\begin{align}
 p(\bm{Y}|\bm{X}) = \prod_{j=1}^{J+1} p(y_j | \bm{y}_{0:j-1}, \bm{X}),
\end{align}
where $y_0$ and $y_{J+1}$ are special tokens representing beginning-of-sentence (BOS) and end-of-sentence (EOS) respectively, $\bm{X} = \bm{x}_{1:I}$, and $\bm{Y} = \bm{y}_{1:J+1}$.

In the training phase, we optimize the parameters $\bm{\theta}$ to minimize the negative log-likelihood in the training data.
Let $\mathcal{D}$ be the training data consisting of a set of pairs of $\bm{X}_n$ and $\bm{Y}_n$: $\mathcal{D} = \{ (\bm{X}_{n}, \bm{Y}_{n}) \}_{n=1}^{|\mathcal{D}|}$.
We minimize the following loss function:
\begin{align}
 \label{eq:loss}
 \mathcal{L}(\bm{\theta}) = -\frac{1}{|\mathcal{D}|} \sum_{(\bm{X}, \bm{Y}) \in \mathcal{D}} \log p(\bm{Y}|\bm{X}; \bm{\theta}).
\end{align}

\section{Definition of Perturbations}\label{sec:perturbation}
This section briefly describes perturbations used in this study.
This study focuses on three types of perturbations: \textit{word replacement}, \textit{word dropout}, and \textit{adversarial perturbations}.
\begin{figure}[!t]
  \centering 
  \includegraphics[width=8cm]{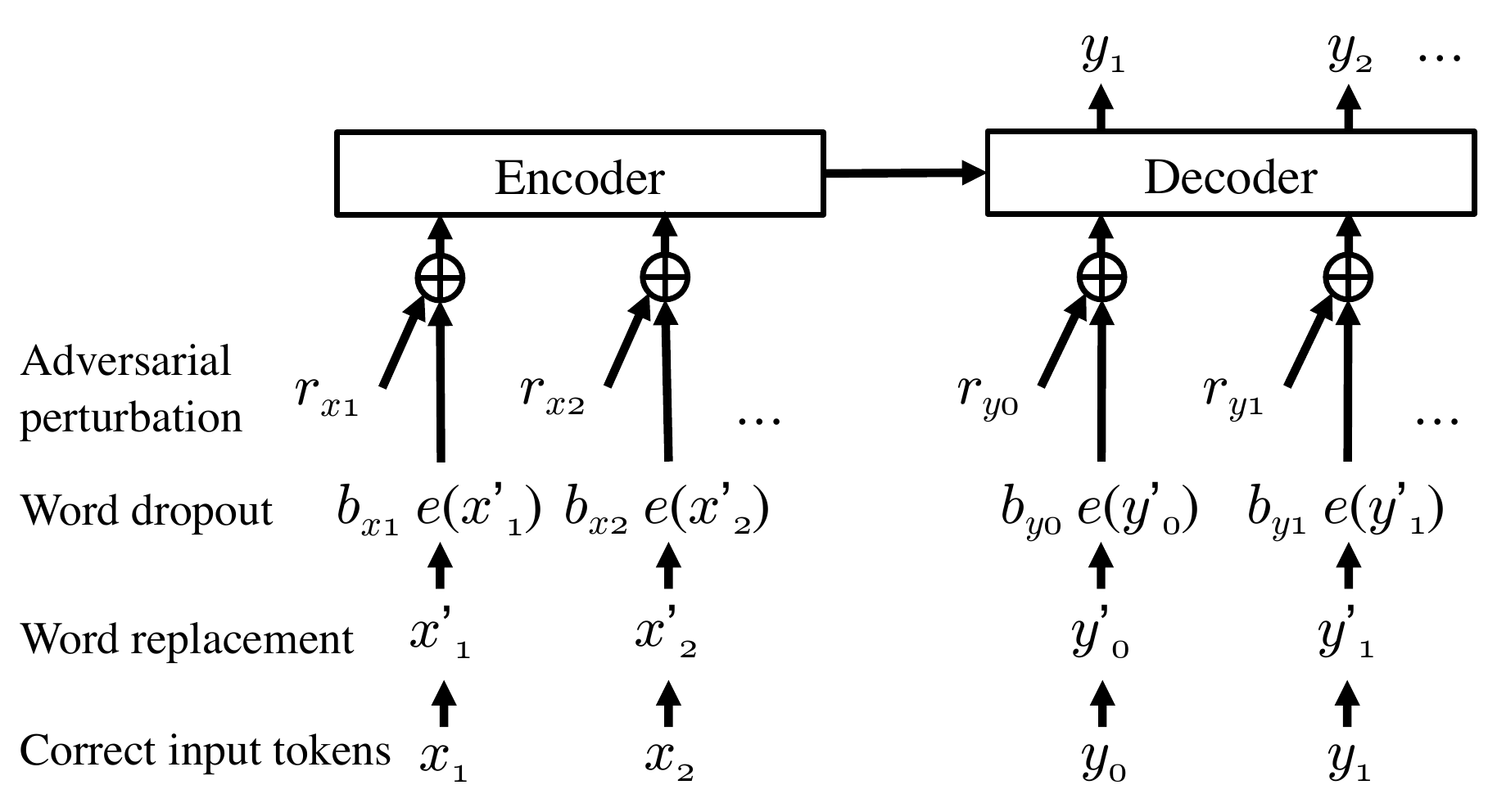}
   \caption{Overview of perturbations used in this study. We can combine perturbations as shown in this figure because each type of perturbation is orthogonal.}
   \label{fig:overview}
\end{figure}
Figure \ref{fig:overview} shows perturbations used in this study.
As shown in this figure, we can use all types of perturbations in the same time because perturbations are orthogonal to each other.
In fact, we combine word replacement with word dropout in our experiments.

\subsection{Word Replacement: \textsc{Rep}}
For any approach that uses a sampled token instead of a correct token, such as the scheduled sampling~\cite{NIPS2015_5956}, we refer to this as a word replacement approach.
In this approach, we construct a new sequence whose tokens are randomly replaced with sampled tokens.
For the construction from the sequence $\bm{X}$, we sample $\hat{x}_{i}$ from a distribution $Q_{x_{i}}$ and use it for the new sequence $\bm{X}'$ with the probability $\alpha$:
\begin{align}
  \hat{x}_i &\sim Q_{x_{i}},\\
 x'_i &= \begin{cases}
   \label{eq:word_replacement}
   x_i  &\textrm{with probability} \ \ \alpha \\
   \hat{x}_i &\textrm{with probability} \ \ 1 - \alpha.
  \end{cases}
\end{align}
We construct $\bm{Y}'$ from the sequence $\bm{Y}$ in the same manner.

\newcite{NIPS2015_5956} used a curriculum learning strategy to adjust $\alpha$, and thus proposed several functions to decrease $\alpha$ based on the training step.
Their strategy uses correct tokens frequently at the beginning of training, whereas it favors sampled tokens frequently at the end of training.
We also adjust $\alpha$ with their use of the inverse sigmoid decay:
\begin{align}
 \alpha_t = \max \left(q, \ \frac{k}{k + \mathrm{exp}({\frac{t}{k}})} \right)
\end{align}
where $q$ and $k$ are hyper-parameters.
In short, $\alpha_t$ decreases to $q$ from $1$, depending on the training step $t$.
We use $\alpha_t$ as $\alpha$ at $t$.

For $Q_{x_{i}}$, we prepare three types of distributions: \textit{conditional probability}, \textit{uniform}, and \textit{similarity}.

\paragraph{Conditional Probability: \parass{}}
\newcite{NIPS2015_5956} proposed the scheduled sampling which uses predicted tokens during training to address the gap between training and inference.
Formally, the scheduled sampling uses the following conditional probability as $Q_{y_{i}}$:
\begin{align}
 \label{eq:shceduled_sampling}
 p(\hat{y}_i | \bm{y'}_{0:i-1}, \bm{X}).
\end{align}
Since the scheduled sampling is the method to compute the perturbation for the decoder side only, it uses the correct sequence as the input of the encoder side.
In other words, the scheduled sampling does not provide any function for $Q_{x_{i}}$.

The original scheduled sampling repeats the decoding for each of the tokens on the decoder side, and thus, requires computational time in proportion to the length of the decoder-side input sequence.
To address this issue, \newcite{DBLP:journals/corr/abs-1906-04331} proposed a more time efficient method: parallel scheduled sampling which computes output probability distributions corresponding to each position simultaneously.
In this study, we use parallel scheduled sampling instead of the original method.

\paragraph{Uniform: \uniform{}}
The scheduled sampling is slow even if we use parallel scheduled sampling because it requires decoding at least once to compute Equation (\ref{eq:shceduled_sampling}).
Thus, we introduce two faster methods to explore effective perturbations from the perspective of computational time.
In \textit{uniform}, we use the uniform distributions on each vocabulary as $Q_{x_{i}}$ and $Q_{y_{i}}$, respectively.
For example, we randomly pick up a token from the source-side vocabulary and use the token as $\hat{x}_i$ in Equation (\ref{eq:word_replacement}) to construct the source-side perturbed input.
This method is used as the baseline in the previous study~\cite{NIPS2015_5956}.

\paragraph{Similarity: \similarity{}}
We also explore more sophisticated way than the uniform distribution.
We assume that the conditional probability of Equation (\ref{eq:shceduled_sampling}) assigns high probabilities to tokens that are similar to the correct input token.
Based on this assumption, we construct a distribution that enables us to sample similar tokens frequently.
Let $\mathcal{V}_x$ be the source-side vocabulary, $\bm{E}_x \in \mathbb{R}^{|\mathcal{V}_x | \times d_x}$ be the $d_x$ dimensional embedding matrix, and $\bm{e}(x_i)$ be the function returning the embedding of $x_i$.
We use the following probability distribution as $Q_{x_{i}}$:
\begin{align}
 \label{eq:similarity}
 \mathrm{softmax}(\bm{E}_x \bm{e}(x_i)),
\end{align}
where $\mathrm{softmax}(.)$ is the softmax function.
Thus, Equation (\ref{eq:similarity}) assigns high probabilities to tokens whose embeddings are similar to $\bm{e}(x_i)$.
In other words, Equation (\ref{eq:similarity}) is the similarity against $x_i$ without considering any context.
We compute the probability distribution for the target side by using $\bm{e}(y_i)$ in the same manner.

\subsection{Word Dropout: \worddrop{}}
We apply the word dropout technique to compute the perturbed input.
Word dropout randomly uses the zero vector instead of the embedding $\bm{e}(x_i)$ for the input token $x_i$~\cite{NIPS2016_6241}:
\begin{align}
 b_{x_i} &\sim \mathrm{Bernoulli}(\beta) , \\
 \mathrm{WDrop}(x_i, b_{x_i}) &= b_{x_i} \bm{e}(x_i) \label{eq:worddrop},
\end{align}
where $\mathrm{Bernoulli}(\beta)$ returns 1 with the probability $\beta$ and 0 otherwise.
Thus, $\mathrm{WDrop}(x_i, b_{x_i})$ returns $\bm{e}(x_i)$ with the probability $\beta$ and the zero vector otherwise.
We apply Equation (\ref{eq:worddrop}) to each token in the input sequence.
Then, we use the results as the perturbed input.

\subsection{Adversarial Perturbation: \adv{}}
\newcite{Miyato2017} proposed a method to compute adversarial perturbations in the embedding space.
Their method adds adversarial perturbations to input embeddings instead of replacing correct input tokens with others.
\newcite{sato-etal-2019-effective} applied this approach to neural encoder-decoders and reported its effectiveness.
Thus, this study follows the methods used in \newcite{sato-etal-2019-effective}.

The method seeks the adversarial perturbation, which seriously damages the loss value, based on the gradient of the loss function $\mathcal{L}(\bm{\theta})$.
Then, we add the adversarial perturbation to the input token embedding.
Let $\bm{r}_{x_i} \in \mathbb{R}^{d_x}$ be the adversarial perturbation vector for the input token $x_i$.
We obtain the perturbed input embedding $\bm{e}'(x_i)$ with the following equations:
\begin{align}
 \bm{e}'(x_i) &= \bm{e}(x_i) + \bm{r}_{x_i}, \\
 \bm{r}_{x_i} &= \epsilon \frac{ \bm{c}_{x_i} }{ ||\bm{c}_{x_i}|| }, \\
 \bm{c}_{x_i} &= \nabla_{\bm{e}({x_i})}\mathcal{L}(\bm{\theta}),
\end{align}
where $\epsilon$ is a hyper-parameter to control the norm of the adversarial perturbation.
We apply the above equations to all tokens in the input sequence.

\subsection{Training}
In the training using word replacement and/or word dropout perturbations, we search the parameters predicting the correct output sequence from the perturbed input.
For example, in the word replacement approach, we minimize the following negative log-likelihood:
\begin{align}
 \label{eq:loss_with_perturbation}
  \mathcal{L}'(\bm{\theta}) &= -\frac{1}{|\mathcal{D}|} \sum_{\mathcal{D}} \log p(\bm{Y}|\bm{X}', \bm{Y}'; \bm{\theta}), \notag \\
   &= -\frac{1}{|\mathcal{D}|} \sum_{\mathcal{D}} \sum_{j=1}^{J+1} \log p(y_j | \bm{y}'_{0:j-1}, \bm{X}'; \bm{\theta}).
\end{align}

\paragraph{Virtual Adversarial Training}
When we use adversarial perturbations, we train parameters of the neural encoder-decoder to minimize both Equation (\ref{eq:loss}) and a loss function $\mathcal{A}(\bm{\theta})$ composed of perturbed inputs:
\begin{align}
 \mathcal{J}(\bm{\theta}) = \mathcal{L}(\bm{\theta}) + \lambda \mathcal{A}(\bm{\theta}),
\end{align}
where $\lambda$ is a hyper-parameter to control the balance of two loss functions.
This calculation seems to be reasonably time efficient because adversarial perturbations require computing Equation (\ref{eq:loss}).

\newcite{sato-etal-2019-effective} used the virtual adversarial training originally proposed in \newcite{miyato2016distributional} as a loss function for perturbed inputs.
In the virtual adversarial training, we regard the output probability distributions given the correct input sequence as positive examples:
\begin{align}
 \mathcal{A}(\bm{\theta}) = \frac{1}{|\mathcal{D}|} \sum_\mathcal{D} \mathrm{KL}\left( p(\cdot | \bm{X}; \bm{\theta}) || p(\cdot | \bm{X}, \bm{r}; \bm{\theta}) \right),
\end{align}
where $\bm{r}$ represents a concatenated vector of adversarial perturbations for each input token, and $\mathrm{KL}(\cdot || \cdot)$ denotes the Kullback–Leibler divergence.

\section{Experiments on Machine Translation}
To obtain findings on sequence-to-sequence problems, we conduct experiments on various situations: different numbers of training data and multiple tasks.
We mainly focus on translation datasets because machine translation is a typical sequence-to-sequence problem.
We regard the widely used WMT English-German dataset as a standard setting.
In addition, we vary the number of training data in machine translation: high resource in Section~\ref{sec:ex_mt_high_resource} and low resource in Section~\ref{sec:ex_mt_low_resource}.
Table \ref{table:mt_dataset} summarizes the number of training data in each configuration.
Moreover, we conduct experiments on other sequence-to-sequence problems: grammatical error correction (GEC) in Section \ref{sec:gec} and summarization in Appendix~\ref{sec:exp_summarization} to confirm whether the findings from machine translation are applicable to other tasks.

\subsection{Standard Setting}
\label{sec:exp_standard_mt}

\begin{table}[!t]
  \centering
  \begin{tabular}{ c | c | c } \hline
  Setting & Genuine & Synthetic \\ \hline
  Standard & 4.5M & - \\
  High Resource & 4.5M & 20.0M \\
  Low Resource & 160K & - \\ \hline
  \end{tabular}
  \caption{Sizes of training datasets on our machine translation experiments.}
  \label{table:mt_dataset}
\end{table}
\paragraph{Datasets}
We used the WMT 2016 English-German training set, which contains 4.5M sentence pairs, in the same as \newcite{ott-etal-2018-scaling}, and followed their pre-processing.
We used newstest2013 as a validation set, and newstest2010-2012, and 2014-2016 as test sets.
We measured case-sensitive detokenized BLEU with SacreBLEU~\cite{post-2018-call}\footnote{As reported in \newcite{ott-etal-2018-scaling}, the BLEU score from SacreBLEU is often lower than the score from \texttt{multi-bleu.perl} but SacreBLEU is suitable for scoring WMT datasets~\cite{post-2018-call}.}.

\paragraph{Methods}
We used Transformer~\cite{NIPS2017_7181} as a base neural encoder-decoder model because it is known as a strong neural encoder-decoder model.
We used two parameter sizes: base and big settings in \newcite{NIPS2017_7181}.

We applied perturbations described in Section \ref{sec:perturbation} for comparison.
For parallel scheduled sampling~\cite{DBLP:journals/corr/abs-1906-04331}, we can compute output probability distributions multiple times but we used the first decoding result only because it is the fastest approach.
We set $q = 0.9$, $k = 1000$, and $\beta = 0.9$.
For \adv{}, we used the same hyper-parameters as in \newcite{sato-etal-2019-effective}.
Our implementation is based on \texttt{fairseq}\footnote{\href{https://github.com/pytorch/fairseq}{https://github.com/pytorch/fairseq}} \cite{ott-etal-2019-fairseq}.
We trained each model for a total of 50,000 steps.

\paragraph{Preliminary: To which sides do we apply perturbations?}
\begin{table*}[!t]
  \centering
  \footnotesize
  \begin{tabular}{ l | c | c c c c c c c | c } \hline
  Method & Position & 2010 & 2011 & 2012 & 2013 & 2014 & 2015 & 2016 & Average \\ \hline \hline
  \multicolumn{10}{c}{Transformer (base)} \\ \hline \hline
  w/o perturbation & - & 24.27 & 22.06 & 22.43 & 26.11 & 27.13 & 29.70 & 34.40 & 26.59 \\ \hline
  \multirow{3}{*}{\uniform{}} & enc & 24.26 & 21.95 & 22.33 & 25.76 & 26.70 & 29.08 & \textbf{34.61} & 26.38 \\
  & dec & 24.27 & 21.99 & 22.29 & \textbf{26.31} & \textbf{27.28} & \textbf{29.74} & \textbf{34.42} & \textbf{26.61} \\
  & both & \textbf{24.30} & \textbf{22.20} & 22.43 & 26.06 & 26.82 & 29.42 & 34.13 & 26.48 \\ \hline
  \multirow{3}{*}{\similarity{}} & enc & 24.12 & 22.02 & 22.14 & \textbf{26.21} & 27.01 & 29.33 & \textbf{34.56} & 26.48\\
  & dec & \textbf{24.32} & 21.96 & \textbf{22.55} & \textbf{26.36} & \textbf{27.23} & \textbf{29.86} & 34.33 & \textbf{26.66} \\
  & both & 23.94 & 21.85 & 22.29 & 25.84 & 26.61 & 29.50 & 34.20 & 26.32 \\ \hline
  \multirow{3}{*}{\worddrop{}} & enc & \textbf{24.31} & \textbf{22.12} & \textbf{22.45} & \textbf{26.20} & 27.09 & \textbf{29.95} & \textbf{34.58} & \textbf{26.67} \\
  & dec & 23.96 & \textbf{22.08} & 22.22 & \textbf{26.36} & 27.08 & \textbf{29.91} & 33.98 & 26.51 \\
  & both & \textbf{24.33} & \textbf{22.14} & 22.35 & 26.10 & 26.82 & 29.51 & \textbf{34.51} & 26.54 \\ \hline \hline
  \multicolumn{10}{c}{Transformer (big)} \\ \hline \hline 
  w/o perturbation & - & 24.22 & 22.11 & 22.69 & 26.60 & 28.46 & 30.50 & 33.58 & 26.88 \\ \hline
  \multirow{3}{*}{\uniform{}} & enc &  \textbf{24.79} & \textbf{22.49} & \textbf{23.10} & \textbf{27.07} & 28.39 & \textbf{30.52} & \textbf{34.51} & \textbf{27.27} \\
  & dec &  \textbf{24.33} & \textbf{22.34} & 22.63 & \textbf{26.93} & 28.22 & 30.36 & 33.41 & \textbf{26.89} \\
  & both & \textbf{24.75} & \textbf{22.68} & \textbf{23.32} & \textbf{27.01} & \textbf{28.89} & \textbf{31.38} & \textbf{34.94} & \textbf{27.57} \\ \hline
  \multirow{3}{*}{\similarity{}} & enc &\textbf{24.68} & \textbf{22.91} & \textbf{23.13} & \textbf{27.03} & 28.25 & \textbf{30.81} & \textbf{34.40} & \textbf{27.32} \\ 
  & dec & \textbf{24.51} & \textbf{22.22} & \textbf{22.83} & 26.46 & \textbf{28.64} & \textbf{30.68} & 33.58 & \textbf{26.99} \\
  & both &\textbf{24.77} & \textbf{22.50} & \textbf{23.10} & \textbf{26.91} & \textbf{28.98} & \textbf{31.03} & \textbf{34.29}& \textbf{27.37} \\ \hline
  \multirow{3}{*}{\worddrop{}} & enc & \textbf{24.60} & \textbf{22.32} & \textbf{23.27} & \textbf{27.07} & 28.40 & \textbf{31.00} & \textbf{34.61} & \textbf{27.32} \\
  & dec & \textbf{24.53} & \textbf{22.33} & \textbf{22.75} & \textbf{27.00} & \textbf{28.56} & \textbf{30.58} & 33.20 & \textbf{26.99} \\
  & both & \textbf{24.92} & \textbf{22.71} & \textbf{23.40} & \textbf{27.11} & \textbf{28.73} & \textbf{30.99} & \textbf{34.80} & \textbf{27.52} \\ \hline
  \uniform{}+\worddrop{} & both & \textbf{24.82} & \textbf{22.82} & \textbf{23.38} & \textbf{27.30} & \textbf{28.56} & \textbf{30.65} & \textbf{35.02} & \textbf{27.51} \\ \hline
  \similarity{}+\worddrop{} & both & \textbf{24.83} & \textbf{22.95} & \textbf{23.40} & \textbf{27.23} & \textbf{28.65} & \textbf{30.88} & \textbf{35.05} & \textbf{27.57} \\ \hline
  \end{tabular}
  \caption{BLEU scores on newstest2010-2016 and averaged scores. Bolds are better scores than w/o perturbations.\label{tab:exp_for_perturbation_pos}}
\end{table*}

As described, perturbations based on \parass{} can be applied to the decoder side only.
\newcite{sato-etal-2019-effective} reported their method was the most effective when they applied their \adv{} to both encoder and decoder sides.
However, we do not have evidence for suitable sides in applying other perturbations.
Thus, we applied \uniform{}, \similarity{}, and \worddrop{} to the encoder side, decoder side, and both as preliminary experiments.

Table \ref{tab:exp_for_perturbation_pos} shows BLEU scores on newstest2010-2016 and averaged scores when we varied the position of the perturbations.
In this table, we indicate better scores than the original Transformer~\cite{NIPS2017_7181} (w/o perturbation) in bold.
This table shows that it is better to apply word replacement (\uniform{} and \similarity{}) to the decoder side in Transformer (base).
For \worddrop{}, applying the encoder side is slightly better than other positions in Transformer (base).
In contrast, applying perturbations to both sides achieved the best averaged BLEU scores for all methods in Transformer (big).
These results imply that it is better to apply to word replacement and/or word dropout to both encoder and decoder sides if we prepare enough parameters for neural encoder-decoders.
Based on these results, we select methods to compare against scheduled sampling (\parass{}) and adversarial perturbations (\adv{}).

Table \ref{tab:exp_for_perturbation_pos} also shows the results when we combined each word replacement with word dropout (\uniform{}+\worddrop{} and \similarity{}+\worddrop{}).
\similarity{}+\worddrop{} slightly outperformed the separated settings.

\paragraph{Results}

\begin{table*}[!t]
  \centering{}
  \footnotesize
  \begin{tabular}{ l | c | c c c c c c c | c | r } \hline
  Method & Position & 2010 & 2011 & 2012 & 2013 & 2014 & 2015 & 2016 & Average & Speed \\ \hline \hline
  \multicolumn{11}{c}{Transformer (base)} \\ \hline \hline
  w/o perturbation & - & 24.27 & 22.06 & 22.43 & 26.11 & 27.13 & 29.70 & 34.40 & 26.59 & $\times$1.00 \\ \hline
  \uniform{} & dec & 24.27 & 21.99 & 22.29 & 26.31 & \textbf{27.28} & 29.74 & 34.42 & 26.61 & $\times$0.99 \\
  \similarity{} & dec & 24.32 & 21.96 & 22.55 & \textbf{26.36} & 27.23 & 29.86 & 34.33 & 26.66 & $\times$0.95 \\
  \worddrop{} & enc & 24.31 & 22.12 & 22.45 & 26.20 & 27.09 & \textbf{29.95} & 34.58 & 26.67 & $\times$1.00 \\
  \parass{} & dec & 24.18 & 22.03 & 22.38 & 26.04 & 27.15 & 29.77 & 34.24 & 26.54 & $\times$0.88 \\
  \adv{} & both & \textbf{24.34} & \textbf{22.19} & \textbf{22.58} & 26.19 & 27.10 & 29.78 & \textbf{34.89} & \textbf{26.72}& $\times$0.44 \\ \hline \hline
  \multicolumn{11}{c}{Transformer (big)} \\ \hline \hline 
  w/o perturbation & - & 24.22 & 22.11 & 22.69 & 26.60 & 28.46 & 30.50 & 33.58 & 26.88 & $\times$0.60 \\ \hline
  \uniform{} & both & 24.75 & 22.68 & 23.32 & 27.01 & 28.89 & \textbf{31.38} & 34.94 & \textbf{27.57} & $\times$0.60 \\
  \similarity{} & both & 24.77 & 22.50 & 23.10 & 26.91 & \textbf{28.98} & 31.03 & 34.29 & 27.37 & $\times$0.55 \\
  \worddrop{} & both & \textbf{24.92} & 22.71 & \textbf{23.40} & 27.11 & 28.73 & 30.99 & 34.80 & 27.52 & $\times$0.60 \\
  \uniform{}+\worddrop{} & both & 24.82 & 22.82 & 23.38 & \textbf{27.30} & 28.56 & 30.65 & 35.02 & 27.51 & $\times$0.60 \\
  \similarity{}+\worddrop{} & both & 24.83 & \textbf{22.95} & \textbf{23.40} & 27.23 & 28.65 & 30.88 & \textbf{35.05} & \textbf{27.57} & $\times$0.55  \\
  \parass{} & dec & 24.44 & 21.97 & 22.74 & 26.77 & 28.44 & 30.83 & 33.71 & 26.99& $\times$0.52 \\
  \adv{} & both & 24.71 & 22.60 & 23.23 & 26.98 & 28.97 & 30.49 & 34.40 & 27.34& $\times$0.20 \\ \hline
  \end{tabular}
  \caption{BLEU scores on newstest2010-2016, averaged scores, and computational speeds based on Transformer (base) w/o perturbation. Scores in bold denote the best result for each set for Transformer (base) and (big) settings.\label{tab:exp_mt_on_standard}}
\end{table*}

\begin{figure*}
   \centering 
   \includegraphics[width=16cm]{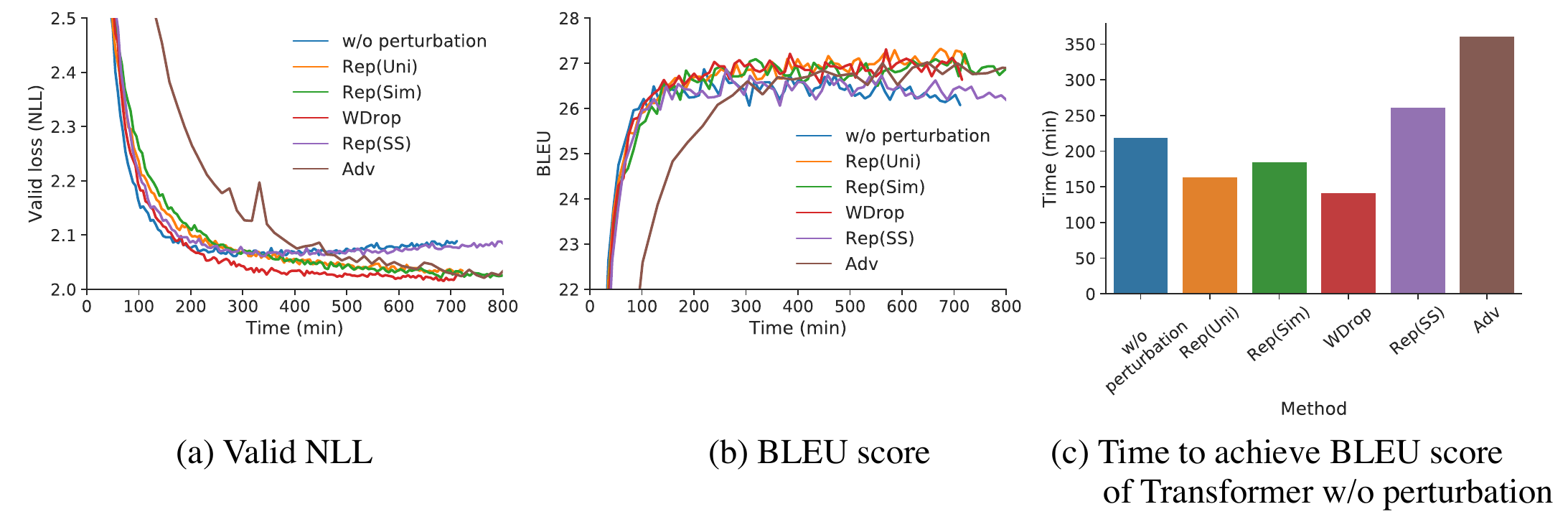}
    \caption{Negative log-likelihood (NLL) values, BLEU scores of each method, and time to achieve BLEU score of Transformer w/o perturbation on validation set (newstest2013).}
    \label{fig:valid_curve}
\end{figure*}

We compare each perturbation in view of computational time.
Table \ref{tab:exp_mt_on_standard} shows BLEU scores of each method and computational speeds\footnote{We regard processed tokens per second as the computational speed of each method.} based on Transformer (base) without any perturbations, i.e., larger is faster.
In this table, we indicate the best score of each column for Transformer (base) and (big) settings in bold.
This table indicates that Transformer without perturbations achieved a comparable score to previous studies~\cite{NIPS2017_7181,ott-etal-2018-scaling} on newstest2014 in base and big settings.
Thus, we consider that our trained Transformer models (w/o perturbation) can be regarded as strong baselines.
This table shows that \adv{} achieved the best averaged score in Transformer (base), but this method required twice as much training time as the original Transformer (base).
In contrast, \similarity{} and \worddrop{} achieved comparable scores to \adv{} although they slightly affected the computational time.
\uniform{} also achieved a slightly better averaged score than the original Transformer (base).

In the Transformer (big) setting, all perturbations surpassed the performance of w/o perturbation in the averaged score.
\parass{} and \adv{} improved the performance, but other methods outperformed these two methods with a small training time.
Moreover, \uniform{} and \similarity{}+\worddrop{} achieved the best averaged score.

Figure \ref{fig:valid_curve} illustrates the negative log-likelihood values and BLEU scores on the validation set for each training time when we applied each perturbation to Transformer (big).
In addition, Figure \ref{fig:valid_curve} (c) shows the time required to achieve the BLEU score of Transformer w/o perturbation on the validation set (26.60, as described in Table \ref{tab:exp_mt_on_standard}).
These figures show that \adv{} requires twice as much time or more relative to other methods to achieve performance comparable to others.
In NLL curves, \uniform{}, \similarity{}, and \worddrop{} achieved better values than those of Transformer w/o perturbation in the early stage.
In addition, \worddrop{} was the fastest to achieve better NLL value.
Figure \ref{fig:valid_curve} (c) indicates that \uniform{}, \similarity{}, and \worddrop{} achieved 26.60 BLEU score with smaller training time than that of Transformer w/o perturbation.

These results indicate that we can quickly improve the performance of Transformer with \uniform{}, \similarity{}, and \worddrop{}.
In particular, when we prepare a large number of parameters for Transformer in machine translation, it is better to use these methods (and their combinations) as perturbations.
We conduct more experiments to investigate whether these methods are also superior in other configurations.

\subsection{High Resource}
\label{sec:ex_mt_high_resource}
\begin{table*}[!t]
  \centering
  \footnotesize
  \begin{tabular}{ l | c | c c c c c c c | c } \hline
  Method & Positions & 2010 & 2011 & 2012 & 2013 & 2014 & 2015 & 2016 & Average \\ \hline
  w/o perturbation & - & 25.63 & 23.62 & 24.54 & 28.39 & 31.50 & 32.96 & 36.47 & 29.02\\ \hline
  \uniform{} & both & 26.36 & 24.18 & 25.14 & 28.54 & \textbf{32.35} & \textbf{33.80} & 37.73 & 29.73 \\
  \similarity{} & both & 26.04 & 23.79 & 25.01 & 28.43 & 32.06 & 33.28 & 37.40 & 29.43 \\
  \worddrop{} & both & \textbf{26.65} & \textbf{24.34} & 25.18 & 28.66 & 32.25 & 33.75 & 37.65 & \textbf{29.78} \\
  \uniform{}+\worddrop{} & both & 26.45 & 24.07 & 25.09 & \textbf{28.72} & 32.21 & 33.42 & 37.68 & 29.66\\
  \similarity{}+\worddrop{} & both & 26.55 & 24.20 & \textbf{25.19} & 28.55 & 31.92 & 33.64 & \textbf{37.96} & 29.72 \\
  \parass{} & dec & 25.81 & 23.64 & 24.73 & 28.46 & 31.84 & 33.29 & 36.59 & 29.19 \\
  \adv{} & both & 25.79 & 24.07 & 24.92 & 28.64 & 32.04 & 33.35 & 37.20 & 29.43 \\ \hline
  \end{tabular}
  \caption{BLEU scores of each method trained with a large amount of data.\label{tab:exp_mt_on_high_resource}}
\end{table*}

We compare each perturbation in the case where we have a large amount of training data.

\paragraph{Datasets}
We add synthetic parallel data generated from the German monolingual corpus using back-translation~\citep{sennrich:2016:backtrans} to the training data used in Section \ref{sec:exp_standard_mt}.
The origin of the German monolingual corpus is NewsCrawl 2015-2018\footnote{\href{http://data.statmt.org/news-crawl/de/}{data.statmt.org/news-crawl/de/}}.
We randomly sampled 5M sentences from each NewsCrawl corpus, and thus, obtained 20M sentences in total.
We back-translated the corpus with the German-English translation model, which is identical to Transformer (big) (w/o perturbation) used in Section~\ref{sec:exp_standard_mt} except for the direction of translation.
Finally, we prepended a special token \tagforbt{} to the beginning of the source (English) side of the synthetic data following \citep{caswell:2019:tagged}.
In addition, we upsampled the original bitext to adjust the ratio of the original and synthetic bitexts to 1:1.

\paragraph{Methods}
In this setting, we increase the parameter size of Transformer from the (big) setting to take advantage of large training data.
Specifically, we increased the internal layer size of the FFN part from 4096 to 8192, and used 8 layers for both the encoder and decoder.
The other hyper-parameters are same as in Section \ref{sec:exp_standard_mt}.

\paragraph{Results}
Table \ref{tab:exp_mt_on_high_resource} shows BLEU scores of each method when we used a large amount of training data.
This table indicates that all perturbations outperformed Transformer w/o perturbation in all test sets.
Moreover, the fast methods \uniform{}, \similarity{}, \worddrop{}, and their combinations achieved the same or better averaged scores than \parass{} and \adv{}.
Thus, these methods are not only fast but also significantly improve the performance of Transformer.
In particular, since Table \ref{tab:exp_mt_on_standard} shows that \uniform{} and \worddrop{} barely have any negative effect on the computational time, we consider them as superior methods.

\subsection{Low Resource}

\paragraph{Datasets}
We also conduct an experiment on a low resource setting.
We used IWSLT 2014 German-English training set which contains 160k sentence pairs.
We followed the preprocessing described in \texttt{fairseq}\footnote{\href{https://github.com/pytorch/fairseq/tree/master/examples/translation}{github.com/pytorch/fairseq/tree/master/examples/translation}} \cite{ott-etal-2019-fairseq}.
We used dev2010, 2012, and tst2010-2012 as a test set.

\paragraph{Methods}
In this setting, we reduced the parameter size of Transformer from the (base) setting.
We reduced the internal layer size of the FFN part from 2048 to 1024.
We used the same values for other hyper-parameters as in Section \ref{sec:exp_standard_mt}.

\paragraph{Results}
\label{sec:ex_mt_low_resource}
\begin{table}[!t]
  \centering
  \footnotesize
  \begin{tabular}{ l | c | r } \hline
  Method & Position & BLEU \\ \hline 
  w/o perturbation & - & 35.22 \\ \hline
  \uniform{} & both & 35.53\\
  \similarity{} & both & 35.49 \\
  \worddrop{} & both & 35.64 \\
  \parass{} & dec & 35.49 \\
  \adv{} & both & \textbf{36.09} \\ \hline
  \multicolumn{3}{c}{$\times$2 training steps} \\ \hline
  \uniform{} & both & 35.81 \\
  \similarity{} & both & 35.96 \\
  \worddrop{} & both & 36.06 \\
  \uniform{}+\worddrop{} & both & 36.20 \\
  \similarity{}+\worddrop{} & both & \textbf{36.22}\\ \hline
  \end{tabular}
  \caption{BLEU scores in the low resource setting.\label{tab:exp_mt_on_low_resource}}
\end{table}

Table \ref{tab:exp_mt_on_low_resource} shows BLEU scores of each method on the low resource setting.
We trained three models with different random seeds for each method, and reported the averaged scores.
In this table, we also report the results of \uniform{}, \similarity{}, \worddrop{}, and their combinations trained with twice the number of updates (below $\times$2 training steps).
This table shows that all perturbations also improved the performance from Transformer w/o perturbation.
In contrast to Tables \ref{tab:exp_mt_on_standard} and \ref{tab:exp_mt_on_high_resource}, \adv{} achieved the top score when each model was trained with the same number of updates.

However, as reported in Section \ref{sec:exp_standard_mt}, \adv{} requires twice or more as long as other perturbations for training.
Thus, when we train Transformer with other perturbations with twice the number of updates, the training time is almost equal.
In the comparison of (almost) equal training time, \worddrop{} achieved a comparable score to \adv{}.
Moreover, \uniform{}+\worddrop{} and \similarity{}+\worddrop{}\footnote{In the low resource setting, we applied only \worddrop{} to an encoder side for \uniform{}+\worddrop{} and \similarity{}+\worddrop{} because the configuration achieved better performance than applying both perturbations to both sides.} outperformed \adv{}.
Thus, in this low resource setting, \uniform{}+\worddrop{} and \similarity{}+\worddrop{} are slightly better than \adv{} in computational time.

\subsection{Results on Perturbed Inputs}
\label{sec:exp_on_perturbed_input}

Recent studies have used perturbations, especially adversarial perturbations, to improve the robustness of encoder-decoders~\cite{sato-etal-2019-effective,cheng-etal-2019-robust,pmlr-v97-wang19f}.
In particular, \newcite{cheng-etal-2019-robust} analyzed the robustness of models trained with their adversarial perturbations over perturbed inputs.
Following them, we also investigate the robustness of our trained Transformer (big) models.

We constructed perturbed inputs by replacing words in source sentences based on pre-defined ratio.
If the ratio is 0.0, we use the original source sentences.
In contrast, if the ratio is 1.0, we use the completely different sentences as source sentences.
We set the ratio $0.01$, $0.05$, and $0.10$.
In this process, we replaced a randomly selected word with a word sampled from vocabulary based on uniform distribution.
We applied this procedure to source sentences in newstest2010-2016.

Table \ref{tab:exp_on_perturbed_input} shows averaged BLEU scores\footnote{For more details, Tables \ref{tab:exp_on_perturbed_0.01}, \ref{tab:exp_on_perturbed_0.05}, and \ref{tab:exp_on_perturbed_0.10} in Appendix show BLEU scores on each perturbed input.} of each method on perturbed newstest2010-2016.
These BLEU scores are calculated against the original reference sentences.
This table indicates that all perturbations improved the robustness of the Transformer (big) because their BLEU scores are better than one in the setting w/o perturbation.
In comparison among perturbations, \similarity{} (and \similarity{}+\worddrop{}) achieved significantly better scores than others on perturbed inputs.
We emphasize that \similarity{} surpassed \adv{} even though \adv{} is originally proposed to improve the robustness of models.
This result implies that \similarity{} is effective to construct robust models as well as to improve the performance.

\begin{table}[!t]
  \centering
  \footnotesize
  \begin{tabular}{ l | c | c | c | c } \hline
  Method & 0.00 & 0.01 & 0.05 & 0.10 \\ \hline 
  w/o perturbation & 26.88 & 25.81 & 21.94 & 17.80 \\ \hline
  \uniform{} & \textbf{27.57} & 26.58 & 23.14 & 18.93 \\
  \similarity{} & 27.37 & 26.95 & 25.34 & 23.13 \\
  \worddrop{} & 27.52 & 26.48 & 22.84 & 18.56 \\
  \uniform{}+\worddrop{} & 27.51 & 26.55 & 23.18 & 19.05\\
  \similarity{}+\worddrop{} & \textbf{27.57} & \textbf{27.15} & \textbf{25.60} & \textbf{23.58} \\
  \parass{} & 26.99 & 25.89 & 22.08 & 17.93 \\
  \adv{} & 27.34 & 26.32 & 22.43 & 18.08 \\ \hline
  \end{tabular}
  \caption{Averaged BLEU scores on newstest2010-2016 when we inject perturbations to a source sentence with each ratio.\label{tab:exp_on_perturbed_input}}
\end{table}

\section{Experiments on GEC}
\label{sec:gec}

\paragraph{Datasets}
Following \newcite{kiyono:2020:ieee}, we used a publicly available dataset from the BEA shared task~\cite{bryant:2019:bea}.
This dataset contains training, validation, and test splits.
We also used the CoNLL-2014 test set (CoNLL)~\citep{ng:2014:conll} as an additional test set.
We report \fscore{} score measured by the ERRANT scorer~\cite{bryant:2017:automatic,felice:2016:automatic} for the BEA dataset and $M^2$ scorer~\cite{dahlmeier:2012:M2} for CoNLL.

\paragraph{Methods}
We used the same settings as \newcite{kiyono:2020:ieee}.
Specifically, we trained Transformer (big) model w/o perturbation on the same parallel pseudo data provided by \newcite{kiyono:2020:ieee}, and then fine-tuned the model with perturbations.
The hyper-parameters for perturbations are the same as those described in Section~\ref{sec:exp_standard_mt}.

\paragraph{Results}
\begin{table}[!t]
  \centering
  \footnotesize
  \begin{tabular}{ l | c | c c c } \hline
  Method & Pos & Valid & Test & CoNLL  \\ \hline
  w/o perturbation & - & 47.25 & 64.74 & 61.62  \\ \hline
  \uniform{} & both & 47.77 & 64.67 & 62.22 \\ 
  \similarity{} & both & 47.58 & 64.51 & 62.29 \\ 
  \worddrop{} & both & 48.53 & 65.47  & 62.22\\ 
  \uniform{}+\worddrop{} & both & 48.58 & 65.94 & \textbf{62.33}\\ 
  \similarity{}+\worddrop{} & both & \textbf{48.72} & \textbf{65.97} & 62.29 \\ 
  \parass{} & dec & 47.84 & 65.18 & 62.30 \\ 
  \adv{} & both & 48.17 & 65.90  & 62.23  \\ \hline
 \newcite{kiyono:2020:ieee} & - & - & 65.0 \ \ & 62.2 \ \ \\ \hline
  \end{tabular}
  \caption{\fscore{} scores of each method. The row of \newcite{kiyono:2020:ieee} represents the reported scores of the model trained with the same configuration.}
  \label{table:gec}
\end{table}

Table \ref{table:gec} shows the results of each method.
This table reports the averaged score of five models trained with different random seeds.
Moreover, we present the scores of \newcite{kiyono:2020:ieee}; our ``w/o perturbation'' model is a rerun of their work, that is, the experimental settings are identical\footnote{The scores of w/o perturbation are slightly worse than \newcite{kiyono:2020:ieee}. We consider that this is due to randomness in the training procedure.}.

Table \ref{table:gec} shows that all perturbations improved the scores except for \uniform{} and \similarity{} in the BEA test set (Test).
Similar to the machine translation results, the simple methods \worddrop{}, \uniform{}+\worddrop{}, and \similarity{}+\worddrop{} achieved comparable scores to \adv{}.
Thus, these faster methods are also effective for the GEC task.

\section{Related Work}
\paragraph{Word Replacement}
The naive training method of neural encoder-decoders has a discrepancy between training and inference; we use the correct tokens as inputs of the decoder in the training phase but use the token predicted at the previous time step as an input of the decoder in the inference phase.
To address this discrepancy, \newcite{NIPS2015_5956} proposed the scheduled sampling that stochastically uses the token sampled from the output probability distribution of the decoder as an input instead of the correct token.
\newcite{zhang-etal-2019-bridging} modified the sampling method to improve the performance.
In addition, \newcite{DBLP:journals/corr/abs-1906-04331} refined the algorithm to be suited to Transformer~\cite{NIPS2017_7181}.
Their method is faster than the original scheduled sampling but slower and slightly worse than more simple replacement methods such as \uniform{} and \similarity{} in our experiments.
\newcite{DBLP:journals/corr/XieWLLNJN17} and \newcite{kobayashi-2018-contextual} used the unigram language model and neural language model respectively to sample tokens for word replacement.
In this study, we ignored contexts to simplify the sampling process, and indicated that such simple methods are effective for sequence-to-sequence problems.

\paragraph{Word Dropout}
\newcite{NIPS2016_6241} applied word dropout to a neural language model and it is a common technique in language modeling~\cite{merityRegOpt,DBLP:journals/corr/abs-1711-03953,D18-1489}.
\newcite{sennrich-zhang-2019-revisiting} reported that word dropout is also effective for low resource machine translation. 
However, word dropout has not been commonly used in the existing sequence-to-sequence systems.
Experiments in this study show that word dropout is not only fast but also contributes to improvement of scores in various sequence-to-sequence problems.

\paragraph{Adversarial Perturbations}
Adversarial perturbations were first discussed in the field of image processing~\cite{42503,Goodfellow2015}.
In the NLP field, \newcite{Miyato2017} applied adversarial perturbations to an embedding space and reported its effectiveness on text classification tasks.
In sequence-to-sequence problems, \newcite{pmlr-v97-wang19f} and \newcite{sato-etal-2019-effective} applied adversarial perturbations to embedding spaces in neural encoder-decoders.
Moreover, \newcite{sato-etal-2019-effective} used virtual adversarial training~\cite{miyato2016distributional} in training of their neural encoder-decoders.
\newcite{cheng-etal-2019-robust} computed token-level adversarial perturbations in machine translation.
In other words, they introduced the strategy of adversarial perturbations into word replacement.
Their method is also effective but requires more computational time than \newcite{pmlr-v97-wang19f} and \newcite{sato-etal-2019-effective} because it runs language models to obtain candidate tokens for perturbations.

\section{Conclusion}
We compared perturbations for neural encoder-decoders in view of computational time.
Experimental results show that simple techniques such as word dropout~\cite{NIPS2016_6241} and random replacement of input tokens achieved comparable scores to sophisticated perturbations: the scheduled sampling~\cite{NIPS2015_5956} and adversarial perturbations~\cite{sato-etal-2019-effective}, even though those simple methods are faster.
In the low resource setting in machine translation, adversarial perturbations achieved high BLEU score but those simple methods also achieved comparable scores to the adversarial perturbations when we spent almost the same time for training.
For the robustness of trained models, \similarity{} is superior to others.
This study indicates that simple methods are sufficiently effective, and thus, we encourage using such simple perturbations as a first step.
In addition, we hope for researchers of perturbations to use the simple perturbations as baselines to make the usefulness of their proposed method clear.

\section*{Acknowledgements}
We thank Motoki Sato for sharing his code with us to compare adversarial perturbations.
We thank Jun Suzuki and Sosuke Kobayashi for their valuable comments.
We thank anonymous reviewers for their useful suggestions.
This work was supported by JSPS KAKENHI Grant Number JP18K18119 and JST ACT-X Grant Number JPMJAX200I.
The first author is supported by Microsoft Research Asia (MSRA) Collaborative Research Program.


\begin{thebibliography}{47}
\expandafter\ifx\csname natexlab\endcsname\relax\def\natexlab#1{#1}\fi

\bibitem[{Barrault et~al.(2019)Barrault, Bojar, Costa-juss{\`a}, Federmann,
  Fishel, Graham, Haddow, Huck, Koehn, Malmasi, Monz, M{\"u}ller, Pal, Post,
  and Zampieri}]{barrault-etal-2019-findings}
Lo{\"\i}c Barrault, Ond{\v{r}}ej Bojar, Marta~R. Costa-juss{\`a}, Christian
  Federmann, Mark Fishel, Yvette Graham, Barry Haddow, Matthias Huck, Philipp
  Koehn, Shervin Malmasi, Christof Monz, Mathias M{\"u}ller, Santanu Pal, Matt
  Post, and Marcos Zampieri. 2019.
\newblock Findings of the 2019 conference on machine translation ({WMT}19).
\newblock In \emph{Proceedings of the Fourth Conference on Machine Translation
  (WMT)}, pages 1--61.

\bibitem[{Bengio et~al.(2015)Bengio, Vinyals, Jaitly, and
  Shazeer}]{NIPS2015_5956}
Samy Bengio, Oriol Vinyals, Navdeep Jaitly, and Noam Shazeer. 2015.
\newblock Scheduled sampling for sequence prediction with recurrent neural
  networks.
\newblock In \emph{Advances in Neural Information Processing Systems 28
  (NIPS)}, pages 1171--1179.

\bibitem[{Bryant et~al.(2019)Bryant, Felice, Andersen, and
  Briscoe}]{bryant:2019:bea}
Christopher Bryant, Mariano Felice, {\O}istein~E. Andersen, and Ted Briscoe.
  2019.
\newblock The {BEA}-2019 shared task on grammatical error correction.
\newblock In \emph{Proceedings of the Fourteenth Workshop on Innovative Use of
  NLP for Building Educational Applications}, pages 52--75.

\bibitem[{Bryant et~al.(2017)Bryant, Felice, and
  Briscoe}]{bryant:2017:automatic}
Christopher Bryant, Mariano Felice, and Ted Briscoe. 2017.
\newblock {Automatic Annotation and Evaluation of Error Types for Grammatical
  Error Correction}.
\newblock In \emph{Proceedings of the 55th Annual Meeting of the Association
  for Computational Linguistics (ACL)}, pages 793--805.

\bibitem[{Caswell et~al.(2019)Caswell, Chelba, and
  Grangier}]{caswell:2019:tagged}
Isaac Caswell, Ciprian Chelba, and David Grangier. 2019.
\newblock Tagged back-translation.
\newblock In \emph{Proceedings of the Fourth Conference on Machine Translation
  (WMT)}, pages 53--63.

\bibitem[{Cheng et~al.(2019)Cheng, Jiang, and
  Macherey}]{cheng-etal-2019-robust}
Yong Cheng, Lu~Jiang, and Wolfgang Macherey. 2019.
\newblock Robust neural machine translation with doubly adversarial inputs.
\newblock In \emph{Proceedings of the 57th Annual Meeting of the Association
  for Computational Linguistics (ACL)}, pages 4324--4333.

\bibitem[{Dahlmeier and Ng(2012)}]{dahlmeier:2012:M2}
Daniel Dahlmeier and Hwee~Tou Ng. 2012.
\newblock {Better Evaluation for Grammatical Error Correction}.
\newblock In \emph{Proceedings of the 2012 Conference of the North American
  Chapter of the Association for Computational Linguistics (NAACL)}, pages
  568--572.

\bibitem[{Dong et~al.(2019)Dong, Yang, Wang, Wei, Liu, Wang, Gao, Zhou, and
  Hon}]{NEURIPS2019_c20bb2d9}
Li~Dong, Nan Yang, Wenhui Wang, Furu Wei, Xiaodong Liu, Yu~Wang, Jianfeng Gao,
  Ming Zhou, and Hsiao-Wuen Hon. 2019.
\newblock Unified language model pre-training for natural language
  understanding and generation.
\newblock In \emph{Advances in Neural Information Processing Systems 32
  (NeurIPS)}, pages 13063--13075.

\bibitem[{Duckworth et~al.(2019)Duckworth, Neelakantan, Goodrich, Kaiser, and
  Bengio}]{DBLP:journals/corr/abs-1906-04331}
Daniel Duckworth, Arvind Neelakantan, Ben Goodrich, Lukasz Kaiser, and Samy
  Bengio. 2019.
\newblock Parallel scheduled sampling.
\newblock \emph{CoRR}, abs/1906.04331.

\bibitem[{Felice et~al.(2016)Felice, Bryant, and
  Briscoe}]{felice:2016:automatic}
Mariano Felice, Christopher Bryant, and Ted Briscoe. 2016.
\newblock {Automatic Extraction of Learner Errors in {ESL} Sentences Using
  Linguistically Enhanced Alignments}.
\newblock In \emph{{Proceedings of the 26th International Conference on
  Computational Linguistics (COLING)}}, pages 825--835.

\bibitem[{Gal and Ghahramani(2016)}]{NIPS2016_6241}
Yarin Gal and Zoubin Ghahramani. 2016.
\newblock A theoretically grounded application of dropout in recurrent neural
  networks.
\newblock In \emph{Advances in Neural Information Processing Systems 29
  (NIPS)}, pages 1019--1027.

\bibitem[{Goodfellow et~al.(2015)Goodfellow, Shlens, and
  Szegedy}]{Goodfellow2015}
Ian Goodfellow, Jonathon Shlens, and Christian Szegedy. 2015.
\newblock Explaining and harnessing adversarial examples.
\newblock In \emph{Proceedings of the 3rd International Conference on Learning
  Representations (ICLR)}.

\bibitem[{Ji et~al.(2017)Ji, Wang, Toutanova, Gong, Truong, and
  Gao}]{ji-etal-2017-nested}
Jianshu Ji, Qinlong Wang, Kristina Toutanova, Yongen Gong, Steven Truong, and
  Jianfeng Gao. 2017.
\newblock A nested attention neural hybrid model for grammatical error
  correction.
\newblock In \emph{Proceedings of the 55th Annual Meeting of the Association
  for Computational Linguistics (ACL)}, pages 753--762.

\bibitem[{Kiyono et~al.(2020)Kiyono, Suzuki, Mizumoto, and
  Inui}]{kiyono:2020:ieee}
Shun Kiyono, Jun Suzuki, Tomoya Mizumoto, and Kentaro Inui. 2020.
\newblock Massive exploration of pseudo data for grammatical error correction.
\newblock \emph{IEEE/ACM Transactions on Audio, Speech, and Language
  Processing}, 28:2134--2145.

\bibitem[{Kobayashi(2018)}]{kobayashi-2018-contextual}
Sosuke Kobayashi. 2018.
\newblock Contextual augmentation: Data augmentation by words with paradigmatic
  relations.
\newblock In \emph{Proceedings of the 2018 Conference of the North {A}merican
  Chapter of the Association for Computational Linguistics: Human Language
  Technologies (NAACL-HLT)}, pages 452--457.

\bibitem[{Li et~al.(2020)Li, Wallace, Shen, Lin, Keutzer, Klein, and
  Gonzalez}]{li2020train}
Zhuohan Li, Eric Wallace, Sheng Shen, Kevin Lin, Kurt Keutzer, Dan Klein, and
  Joseph~E. Gonzalez. 2020.
\newblock Train large, then compress: Rethinking model size for efficient
  training and inference of transformers.
\newblock In \emph{Proceedings of the 37th International Conference on Machine
  Learning (ICML)}, pages 11432--11442.

\bibitem[{Merity et~al.(2018)Merity, Keskar, and Socher}]{merityRegOpt}
Stephen Merity, Nitish~Shirish Keskar, and Richard Socher. 2018.
\newblock {Regularizing and Optimizing LSTM Language Models}.
\newblock In \emph{Proceedings of the 6th International Conference on Learning
  Representations (ICLR)}.

\bibitem[{Miyato et~al.(2017)Miyato, Dai, and Goodfellow}]{Miyato2017}
Takeru Miyato, Andrew~M. Dai, and Ian Goodfellow. 2017.
\newblock Adversarial training methods for semi-supervised text classification.
\newblock In \emph{Proceedings of the 5th International Conference on Learning
  Representations (ICLR)}.

\bibitem[{Miyato et~al.(2016)Miyato, ichi Maeda, Koyama, Nakae, and
  Ishii}]{miyato2016distributional}
Takeru Miyato, Shin ichi Maeda, Masanori Koyama, Ken Nakae, and Shin Ishii.
  2016.
\newblock Distributional smoothing with virtual adversarial training.
\newblock In \emph{Proceedings of the 4th International Conference on Learning
  Representations (ICLR)}.

\bibitem[{Napoles et~al.(2012)Napoles, Gormley, and
  Van~Durme}]{napoles:2012:AG}
Courtney Napoles, Matthew Gormley, and Benjamin Van~Durme. 2012.
\newblock {Annotated Gigaword}.
\newblock In \emph{Proceedings of the Joint Workshop on Automatic Knowledge
  Base Construction and Web-scale Knowledge Extraction (AKBC-WEKEX)}, pages
  95--100.

\bibitem[{Ng et~al.(2014)Ng, Wu, Briscoe, Hadiwinoto, Susanto, and
  Bryant}]{ng:2014:conll}
Hwee~Tou Ng, Siew~Mei Wu, Ted Briscoe, Christian Hadiwinoto, Raymond~Hendy
  Susanto, and Christopher Bryant. 2014.
\newblock {The CoNLL-2014 Shared Task on Grammatical Error Correction}.
\newblock In \emph{Proceedings of the Eighteenth Conference on Computational
  Natural Language Learning (CoNLL)}, pages 1--14.

\bibitem[{Ott et~al.(2019)Ott, Edunov, Baevski, Fan, Gross, Ng, Grangier, and
  Auli}]{ott-etal-2019-fairseq}
Myle Ott, Sergey Edunov, Alexei Baevski, Angela Fan, Sam Gross, Nathan Ng,
  David Grangier, and Michael Auli. 2019.
\newblock fairseq: A fast, extensible toolkit for sequence modeling.
\newblock In \emph{Proceedings of the 2019 Conference of the North {A}merican
  Chapter of the Association for Computational Linguistics: Human Language
  Technologies (NAACL-HLT)}, pages 48--53.

\bibitem[{Ott et~al.(2018)Ott, Edunov, Grangier, and
  Auli}]{ott-etal-2018-scaling}
Myle Ott, Sergey Edunov, David Grangier, and Michael Auli. 2018.
\newblock Scaling neural machine translation.
\newblock In \emph{Proceedings of the Third Conference on Machine Translation
  (WMT)}, pages 1--9.

\bibitem[{Over et~al.(2007)Over, Dang, and
  Harman}]{Over:2007:DC:1284916.1285157}
Paul Over, Hoa Dang, and Donna Harman. 2007.
\newblock Duc in context.
\newblock \emph{Information Processing \& Management}, 43(6):1506--1520.

\bibitem[{Post(2018)}]{post-2018-call}
Matt Post. 2018.
\newblock A call for clarity in reporting {BLEU} scores.
\newblock In \emph{Proceedings of the Third Conference on Machine Translation
  (WMT)}, pages 186--191.

\bibitem[{Qi et~al.(2020)Qi, Yan, Gong, Liu, Duan, Chen, Zhang, and
  Zhou}]{qi2020prophetnet}
Weizhen Qi, Yu~Yan, Yeyun Gong, Dayiheng Liu, Nan Duan, Jiusheng Chen, Ruofei
  Zhang, and Ming Zhou. 2020.
\newblock {P}rophet{N}et: Predicting future n-gram for
  sequence-to-{S}equence{P}re-training.
\newblock In \emph{Findings of the Association for Computational Linguistics:
  EMNLP 2020}, pages 2401--2410.

\bibitem[{Rush et~al.(2015)Rush, Chopra, and
  Weston}]{rush-chopra-weston:2015:EMNLP}
Alexander~M. Rush, Sumit Chopra, and Jason Weston. 2015.
\newblock {A Neural Attention Model for Abstractive Sentence Summarization}.
\newblock In \emph{Proceedings of the 2015 Conference on Empirical Methods in
  Natural Language Processing (EMNLP)}, pages 379--389.

\bibitem[{Sato et~al.(2019)Sato, Suzuki, and Kiyono}]{sato-etal-2019-effective}
Motoki Sato, Jun Suzuki, and Shun Kiyono. 2019.
\newblock Effective adversarial regularization for neural machine translation.
\newblock In \emph{Proceedings of the 57th Annual Meeting of the Association
  for Computational Linguistics (ACL)}, pages 204--210.

\bibitem[{Schwartz et~al.(2019)Schwartz, Dodge, Smith, and
  Etzioni}]{DBLP:journals/corr/abs-1907-10597}
Roy Schwartz, Jesse Dodge, Noah~A. Smith, and Oren Etzioni. 2019.
\newblock Green {AI}.
\newblock \emph{CoRR}, abs/1907.10597.

\bibitem[{Sennrich et~al.(2016{\natexlab{a}})Sennrich, Haddow, and
  Birch}]{sennrich:2016:backtrans}
Rico Sennrich, Barry Haddow, and Alexandra Birch. 2016{\natexlab{a}}.
\newblock Improving neural machine translation models with monolingual data.
\newblock In \emph{Proceedings of the 54th Annual Meeting of the Association
  for Computational Linguistics (ACL)}, pages 86--96.

\bibitem[{Sennrich et~al.(2016{\natexlab{b}})Sennrich, Haddow, and
  Birch}]{sennrich-etal-2016-neural}
Rico Sennrich, Barry Haddow, and Alexandra Birch. 2016{\natexlab{b}}.
\newblock Neural machine translation of rare words with subword units.
\newblock In \emph{Proceedings of the 54th Annual Meeting of the Association
  for Computational Linguistics (ACL)}, pages 1715--1725.

\bibitem[{Sennrich and Zhang(2019)}]{sennrich-zhang-2019-revisiting}
Rico Sennrich and Biao Zhang. 2019.
\newblock Revisiting low-resource neural machine translation: A case study.
\newblock In \emph{Proceedings of the 57th Annual Meeting of the Association
  for Computational Linguistics (ACL)}, pages 211--221.

\bibitem[{Song et~al.(2019)Song, Tan, Qin, Lu, and Liu}]{song2019mass}
Kaitao Song, Xu~Tan, Tao Qin, Jianfeng Lu, and Tie-Yan Liu. 2019.
\newblock Mass: Masked sequence to sequence pre-training for language
  generation.
\newblock In \emph{International Conference on Machine Learning (ICML)}, pages
  5926--5936.

\bibitem[{Strubell et~al.(2019)Strubell, Ganesh, and
  McCallum}]{strubell-etal-2019-energy}
Emma Strubell, Ananya Ganesh, and Andrew McCallum. 2019.
\newblock Energy and policy considerations for deep learning in {NLP}.
\newblock In \emph{Proceedings of the 57th Annual Meeting of the Association
  for Computational Linguistics (ACL)}, pages 3645--3650.

\bibitem[{Sutskever et~al.(2014)Sutskever, Vinyals, and
  Le}]{Sutskever:2014:SSL:2969033.2969173}
Ilya Sutskever, Oriol Vinyals, and Quoc~V. Le. 2014.
\newblock {Sequence to Sequence Learning with Neural Networks}.
\newblock In \emph{Advances in Neural Information Processing Systems 27
  (NIPS)}, pages 3104--3112.

\bibitem[{Suzuki and Nagata(2017)}]{suzuki-nagata-2017-cutting}
Jun Suzuki and Masaaki Nagata. 2017.
\newblock Cutting-off redundant repeating generations for neural abstractive
  summarization.
\newblock In \emph{Proceedings of the 15th Conference of the {E}uropean Chapter
  of the Association for Computational Linguistics (EACL)}, pages 291--297.

\bibitem[{Szegedy et~al.(2014)Szegedy, Zaremba, Sutskever, Bruna, Erhan,
  Goodfellow, and Fergus}]{42503}
Christian Szegedy, Wojciech Zaremba, Ilya Sutskever, Joan Bruna, Dumitru Erhan,
  Ian Goodfellow, and Rob Fergus. 2014.
\newblock Intriguing properties of neural networks.
\newblock In \emph{Proceedings of the 2nd International Conference on Learning
  Representations (ICLR)}.

\bibitem[{Takase and Kobayashi(2020)}]{takase2020word}
Sho Takase and Sosuke Kobayashi. 2020.
\newblock All word embeddings from one embedding.
\newblock In \emph{Advances in Neural Information Processing Systems 33
  (NeurIPS)}, pages 3775--3785.

\bibitem[{Takase and Okazaki(2019)}]{takase-okazaki-2019-positional}
Sho Takase and Naoaki Okazaki. 2019.
\newblock Positional encoding to control output sequence length.
\newblock In \emph{Proceedings of the 2019 Conference of the North {A}merican
  Chapter of the Association for Computational Linguistics (NAACL-HLT)}, pages
  3999--4004.

\bibitem[{Takase et~al.(2018)Takase, Suzuki, and Nagata}]{D18-1489}
Sho Takase, Jun Suzuki, and Masaaki Nagata. 2018.
\newblock Direct output connection for a high-rank language model.
\newblock In \emph{Proceedings of the 2018 Conference on Empirical Methods in
  Natural Language Processing (EMNLP)}, pages 4599--4609.

\bibitem[{Vaswani et~al.(2017)Vaswani, Shazeer, Parmar, Uszkoreit, Jones,
  Gomez, Kaiser, and Polosukhin}]{NIPS2017_7181}
Ashish Vaswani, Noam Shazeer, Niki Parmar, Jakob Uszkoreit, Llion Jones,
  Aidan~N Gomez, \L~ukasz Kaiser, and Illia Polosukhin. 2017.
\newblock Attention is all you need.
\newblock In \emph{Advances in Neural Information Processing Systems 30
  (NIPS)}, pages 5998--6008.

\bibitem[{Wang et~al.(2019)Wang, Gong, and Liu}]{pmlr-v97-wang19f}
Dilin Wang, Chengyue Gong, and Qiang Liu. 2019.
\newblock Improving neural language modeling via adversarial training.
\newblock In \emph{Proceedings of the 36th International Conference on Machine
  Learning (ICML)}, pages 6555--6565.

\bibitem[{Xie et~al.(2017)Xie, Wang, Li, Levy, Nie, Jurafsky, and
  Ng}]{DBLP:journals/corr/XieWLLNJN17}
Ziang Xie, Sida~I. Wang, Jiwei Li, Daniel Levy, Aiming Nie, Dan Jurafsky, and
  Andrew~Y. Ng. 2017.
\newblock Data noising as smoothing in neural network language models.
\newblock In \emph{Proceedings of the 5th International Conference on Learning
  Representations (ICLR)}.

\bibitem[{Yang et~al.(2018)Yang, Dai, Salakhutdinov, and
  Cohen}]{DBLP:journals/corr/abs-1711-03953}
Zhilin Yang, Zihang Dai, Ruslan Salakhutdinov, and William~W. Cohen. 2018.
\newblock Breaking the softmax bottleneck: {A} high-rank {RNN} language model.
\newblock In \emph{Proceedings of the 6th International Conference on Learning
  Representations (ICLR)}.

\bibitem[{Zellers et~al.(2019)Zellers, Holtzman, Rashkin, Bisk, Farhadi,
  Roesner, and Choi}]{NEURIPS2019_3e9f0fc9}
Rowan Zellers, Ari Holtzman, Hannah Rashkin, Yonatan Bisk, Ali Farhadi,
  Franziska Roesner, and Yejin Choi. 2019.
\newblock Defending against neural fake news.
\newblock In \emph{Advances in Neural Information Processing Systems 32
  (NeurIPS)}, pages 9054--9065.

\bibitem[{Zhang et~al.(2020)Zhang, Zhao, Saleh, and Liu}]{zhang2019pegasus}
Jingqing Zhang, Yao Zhao, Mohammad Saleh, and Peter~J. Liu. 2020.
\newblock Pegasus: Pre-training with extracted gap-sentences for abstractive
  summarization.
\newblock In \emph{Proceedings of the 37th International Conference on Machine
  Learning (ICML)}.

\bibitem[{Zhang et~al.(2019)Zhang, Feng, Meng, You, and
  Liu}]{zhang-etal-2019-bridging}
Wen Zhang, Yang Feng, Fandong Meng, Di~You, and Qun Liu. 2019.
\newblock Bridging the gap between training and inference for neural machine
  translation.
\newblock In \emph{Proceedings of the 57th Annual Meeting of the Association
  for Computational Linguistics (ACL)}, pages 4334--4343.

\end{thebibliography}
\bibliographystyle{acl_natbib}

\clearpage

\appendix

\section{Experiments on Summarization}
\label{sec:exp_summarization}

We conduct experiments on two summarization datasets: Annotated English Gigaword~\cite{napoles:2012:AG,rush-chopra-weston:2015:EMNLP} and DUC 2004 task 1~\cite{Over:2007:DC:1284916.1285157}.

\subsection{Annotated English Gigaword}
\label{sec:summarization_engiga}
\paragraph{Datasets}
We used sentence-summary pairs extracted from Annotated English Gigaword~\cite{napoles:2012:AG,rush-chopra-weston:2015:EMNLP} as the summarization dataset.
This dataset contains 3.8M sentence-summary pairs as the training set and 1951 pairs as the test set.
We extracted 3K pairs from the original validation set, which contains 190K pairs, for our validation set.

In summarization, most recent studies used large scale corpora to pre-train their neural encoder-decoder~\cite{NEURIPS2019_c20bb2d9,song2019mass,zhang2019pegasus,qi2020prophetnet}.
Thus, we also augmented the training data.
We extracted the first sentence and headline of a news article in REALNEWS~\cite{NEURIPS2019_3e9f0fc9} and News Crawl~\cite{barrault-etal-2019-findings} as sentence-summary pairs.
In total, we used 17.1M sentence-summary pairs as our training data.

We used BPE~\cite{sennrich-etal-2016-neural} to construct a vocabulary set.
We set the number of BPE merge operations at 32K and shared the vocabulary between both the encoder and decoder sides.

\paragraph{Methods}
We followed the configuration in Section \ref{sec:ex_mt_high_resource} because it seems suitable for a large amount of training data.
We used the same perturbations and hyper-parameters as in Section \ref{sec:ex_mt_high_resource}.

\paragraph{Results}
Table \ref{tab:exp_engiga} shows the ROUGE F$_1$ scores of each method and scores reported in recent studies~\cite{NEURIPS2019_c20bb2d9,song2019mass,zhang2019pegasus,qi2020prophetnet}
In this experiment, we cannot report the result of \adv{} because the loss value of \adv{} exploded during training.
We tried several random seeds for \adv{}, but all models failed to converge.
Since we need a huge amount of budget to search more suitable hyper-parameters for \adv{} in this summarization dataset, we consider that it is impractical to report the result of \adv{}.

Table \ref{tab:exp_engiga} indicates that all perturbations improved the ROUGE score.
In addition, \uniform{}, \similarity{}, \worddrop{}, and their combinations outperformed the scheduled sampling.
Thus, these fast methods are also superior perturbations in the summarization task.
Moreover, \uniform{} and \worddrop{} outperformed the current top score~\cite{qi2020prophetnet} in ROUGE-1, L and ROUGE-2 respectively.

\begin{table}[!t]
  \centering
  \footnotesize
  \begin{tabular}{ l | c | c c c } \hline
  Method & Position & R-1 & R-2 & R-L \\ \hline
  w/o perturbation & -  & 39.20 & 19.84 & 36.21 \\ \hline
  \uniform{} & both & \textbf{39.81} & 20.40 & \textbf{36.93}\\
  \similarity{} & both & 39.70 & 20.14 & 36.77 \\
  \worddrop{} & both & 39.66 & \textbf{20.45} & 36.59 \\
  \uniform{}+\worddrop{} & both &39.36 & 20.13 & 36.62 \\
  \similarity{}+\worddrop{} & both &39.56 & 20.14 & 36.66 \\
  \parass{} & dec & 39.20 & 20.04 & 36.27 \\ \hline
  \newcite{NEURIPS2019_c20bb2d9} & - & 38.45 & 19.45 & 35.75 \\
  \newcite{song2019mass} & - & 38.73 & 19.71 & 35.96 \\
  \newcite{zhang2019pegasus} & - & 39.12 & 19.86 & 36.24 \\
  \newcite{qi2020prophetnet} & - & 39.51 & 20.42 & 36.69 \\\hline
  \end{tabular}
  \caption{F$_1$ values of ROUGE-1, 2, and L (R-1, R-2, and R-L respectively) on the test set extracted from Annotated English Gigaword. The lower part represents the scores reported by recent studies.\label{tab:exp_engiga}}
\end{table}

\begin{table*}[!t]
  \centering
  \begin{tabular}{ l | c | c c c } \hline
  Method & Position & R-1 & R-2 & R-L \\ \hline
  w/o perturbation & - & 32.80 & 11.55 & 28.26\\ \hline
  \uniform{} & both & 32.56 & 11.48 & 28.21 \\
  \similarity{} & both & 32.80 & 11.55 & 28.28 \\
  \worddrop{} & both & \textbf{33.06} & 11.45 & \textbf{28.51} \\
  \uniform{}+\worddrop{} & both & 32.15 & 11.58 & 28.01\\
  \similarity{}+\worddrop{} & both & 32.80 & \textbf{11.73} & 28.46\\
  \parass & dec & 32.83 & 11.41 & 28.14 \\ \hline
  \newcite{rush-chopra-weston:2015:EMNLP} & - & 28.18 &  \ \ 8.49 & 23.81 \\
  \newcite{suzuki-nagata-2017-cutting} & - & 32.28 & 10.54 & 27.80 \\
  \newcite{takase-okazaki-2019-positional} & - & 32.29 & 11.49 & 28.03 \\
  \newcite{takase2020word} & - & 32.57 & 11.63 & 28.24 \\ \hline
  \end{tabular}
  \caption{Recall-based ROUGE-1, 2, and L (R-1, R-2, and R-L respectively) on DUC 2004 task 1 test set. The lower part represents the scores reported by recent studies.\label{tab:exp_duc}}
\end{table*}

\subsection{DUC 2004 Task 1}

\paragraph{Datasets}
We used sentence-summary pairs extracted from Annotated English Gigaword~\cite{napoles:2012:AG,rush-chopra-weston:2015:EMNLP} and News Crawl~\cite{barrault-etal-2019-findings} as our training dataset, which contains 10.1M pairs.
Following recent studies~\cite{takase-okazaki-2019-positional,takase2020word}, we used BPE to construct a vocabulary set for the encoder side and characters as vocabulary for the decoder side.
We set the number of BPE merge operations at 16K.

For the test set, we used the DUC 2004 task 1 dataset~\cite{Over:2007:DC:1284916.1285157} which contains 500 source sentences and four kinds of manually constructed reference summaries.
We truncated characters over 75 bytes in each generated summary based on the official configuration.

\paragraph{Methods}
We used the Transformer (big) setting in this experiment.
In addition, we introduced the output length control method proposed by \newcite{takase-okazaki-2019-positional}.
We used the same perturbations and hyper-parameters as in Section \ref{sec:exp_standard_mt}.

\paragraph{Results}
Table \ref{tab:exp_engiga} shows recall-based ROUGE scores of each method and scores reported in recent studies~\cite{rush-chopra-weston:2015:EMNLP,suzuki-nagata-2017-cutting,takase-okazaki-2019-positional,takase2020word}.
We also cannot report the result of \adv{} for the same reason as described in Appendix \ref{sec:summarization_engiga}.

This table indicates that \similarity{}, \worddrop{}, and their combination improved the ROUGE scores.
In particular, \similarity{}+\worddrop{} outperformed the current top score in ROUGE-1, 2, and L.
Moreover, \worddrop{} achieved better ROUGE-1 and L scores than the current top score.
In contrast, \uniform{} slightly harmed the performance in this configuration.
These results indicate that \worddrop{} and \similarity{} are also effective for summarization tasks.

\begin{table*}[!t]
  \centering
  \begin{tabular}{ l | c | c c c c c c c | c } \hline
  Method & Positions & 2010 & 2011 & 2012 & 2013 & 2014 & 2015 & 2016 & Average \\ \hline
  w/o perturbation & - & 23.37 & 21.23 & 21.89 & 25.56 & 26.97 & 29.34 & 32.32 & 25.81 \\ \hline
  \uniform{} & both & 23.88 & 21.85 & 22.48 & 26.23 & 27.81 & 30.21 & 33.61 & 26.58 \\
  \similarity{} & both & \textbf{24.51} & 22.24 & 22.82 & 26.53 & \textbf{28.44} & \textbf{30.43} & 33.68 & 26.95 \\
  \worddrop{} & both & 24.01 & 21.90 & 22.48 & 26.24 & 27.60 & 29.71 & 33.44 & 26.48 \\
  \uniform{}+\worddrop{} & both & 23.85 & 22.03 & 22.69 & 26.63 & 27.50 & 29.56 & 33.60 & 26.55 \\
  \similarity{}+\worddrop{} & both & 24.47 & \textbf{22.61} & \textbf{23.15} & \textbf{26.88} & 28.24 & 30.14 & \textbf{34.53} & \textbf{27.15} \\
  \parass{} & dec & 23.49 & 21.18 & 21.82 & 25.79 & 27.17 & 29.39 & 32.38 & 25.89 \\
  \adv{} & both & 23.94 & 21.70 & 22.46 & 25.99 & 27.71 & 29.28 & 33.13 & 26.32 \\ \hline
  \end{tabular}
  \caption{BLEU scores when we inject perturbations to a source sentence with 0.01.\label{tab:exp_on_perturbed_0.01}}
\end{table*}

\begin{table*}[!t]
  \centering
  \begin{tabular}{ l | c | c c c c c c c | c } \hline
  Method & Positions & 2010 & 2011 & 2012 & 2013 & 2014 & 2015 & 2016 & Average \\ \hline
  w/o perturbation & - & 19.78 & 18.51 & 18.70 & 21.58 & 22.74 & 24.81 & 27.45 & 21.94 \\ \hline
  \uniform{} & both & 21.02 & 19.38 & 19.67 & 22.76 & 23.85 & 26.08 & 29.24 & 23.14 \\
  \similarity{} & both & 23.13 & 21.13 & 21.60 & 24.98 & \textbf{26.69} & 28.38 & 31.49 & 25.34 \\
  \worddrop{} & both & 20.94 & 19.24 & 19.44 & 22.41 & 23.67 & 25.42 & 28.74 & 22.84 \\
  \uniform{}+\worddrop{} & both & 20.99 & 19.64 & 19.93 & 22.89 & 23.77 & 25.64 & 29.40 & 23.18 \\
  \similarity{}+\worddrop{} & both & \textbf{23.20} & \textbf{21.55} & \textbf{21.87} & \textbf{25.53} & 26.50 & \textbf{28.49} & \textbf{32.05} & \textbf{25.60} \\
  \parass{} & dec & 20.06 & 18.58 & 18.90 & 21.92 & 23.01 & 24.59 & 27.51 & 22.08 \\
  \adv{} & both & 20.45 & 18.91 & 19.10 & 22.02 & 23.50 & 24.97 & 28.05 & 22.43 \\ \hline
  \end{tabular}
  \caption{BLEU scores when we inject perturbations to a source sentence with 0.05.\label{tab:exp_on_perturbed_0.05}}
\end{table*}

\begin{table*}[!t]
  \centering
  \begin{tabular}{ l | c | c c c c c c c | c } \hline
  Method & Positions & 2010 & 2011 & 2012 & 2013 & 2014 & 2015 & 2016 & Average \\ \hline
  w/o perturbation & - & 16.21 & 15.03 & 15.31 & 17.82 & 17.76 & 19.91 & 22.57 & 17.80 \\ \hline
  \uniform{} & both & 17.24 & 15.84 & 16.39 & 18.62 & 19.30 & 21.45 & 23.64 & 18.93 \\
  \similarity{} & both & 21.15 & 19.18 & 19.79 & 22.95 & 23.91 & 26.19 & 28.73 & 23.13 \\
  \worddrop{} & both & 16.79 & 15.54 & 16.06 & 18.35 & 18.68 & 20.57 & 23.96 & 18.56 \\
  \uniform{}+\worddrop{} & both & 17.53 & 16.00 & 16.41 & 18.95 & 19.40 & 21.03 & 24.01 & 19.05 \\
  \similarity{}+\worddrop{} & both & \textbf{21.58} & \textbf{19.86} & \textbf{20.10} & \textbf{23.50} & \textbf{24.22} & \textbf{26.27} & \textbf{29.55} & \textbf{23.58} \\
  \parass{} & dec & 16.31 & 15.21 & 15.18 & 18.01 & 18.11 & 20.00 & 22.69 & 17.93 \\
  \adv{} & both & 16.47 & 15.24 & 15.50 & 18.01 & 18.07 & 19.84 & 23.44 & 18.08 \\ \hline
  \end{tabular}
  \caption{BLEU scores when we inject perturbations to a source sentence with 0.10.\label{tab:exp_on_perturbed_0.10}}
\end{table*}

\end{document}